\documentclass{article}
\usepackage[utf8]{inputenc}
\usepackage[english]{babel}

\usepackage{amsmath}
\usepackage{graphicx}
\usepackage{caption}
\usepackage{subcaption} 
\usepackage[normalem]{ulem}
\usepackage{tabularray}
\usepackage{enumitem}
\usepackage{pdflscape}
\usepackage{tabularx}
\usepackage[colorlinks=true, allcolors=blue]{hyperref}
\usepackage[table]{xcolor}

 \DefTblrTemplate{capcont}{default}{}
\title{BERT-based Authorship Attribution on the Romanian Dataset called ROST}
\author{Sanda-Maria Avram}
\date{January 2023}

\begin{document}

\maketitle
\begin{abstract}
    Being around for decades, the problem of Authorship Attribution is still very much in focus currently. Some of the more recent instruments used are the pre-trained language models, the most prevalent being BERT. Here we used such a model to detect the authorship of texts written in the Romanian language. The dataset used is highly unbalanced, i.e., significant differences in the  number of texts per author, the sources from which the texts were collected, the time period in which the authors lived and wrote these texts, the medium intended to be read (i.e., paper or online), and the type of writing (i.e., stories, short stories, fairy tales, novels, literary articles, and sketches). The results are better than expected, sometimes exceeding 87\% macro-accuracy. 
\end{abstract}

\section{Introduction}

The problem of automated Authorship Attribution (AA) is not new, however it is still relevant nowadays. It is defined as the task of determining authorship of an unknown text based on the textual characteristics of the text itself~\cite{oliveira2013comparing}.

Most approaches that treat the AA problem are in the area of artificial intelligence and use a simple classifier (e.g., linear SVM or decision tree) having as features bag-of-words (character n-grams) or other conventional feature sets~\cite{kestemont2021overview, kestemont2018overview}. The adoption of using deep neural learning, which was already used for Natural Language Processing (NLP), occurred later for authorship
identification. More recently, pre-trained language models (such as BERT and GPT-2) have been used for finetuning and accuracy improvements~\cite{kestemont2021overview, tyo2022state, barlas2021transfer}.

The paper is organized as follows:
\begin{description}
    \item [Section~\ref{related_work}] describes the related work, more specific instruments that were used in the authorship attribution problem by using methods from artificial intelligence;
    \item [Section~\ref{prerequisite}] presents details pertaining to preparing the considered dataset to be processed by using a Romanian pre-trained BERT model;
    \item [Section~\ref{tests}] provides implementation details and results obtained by the considered BERT model;
    \item [Section~\ref{conclusion}] concludes with final remarks on the work and provides future possible directions and investigations.
\end{description}

\section{Related work}
\label{related_work}

The strategies used for addressing the AA problem are multiple. Authors of~\cite{tyo2022state} grouped the main approaches into 4 classes:
 \begin{description}
   \item [Ngram] -- includes character n-grams, parts-of-speech and summary statistics as shown in~\cite{altakrori-etal-2021-topic-confusion, murauer-specht-2021-developing, bischoff2020importance, stamatatos2018masking};
   \item [PPM] -- uses Prediction by Partial Matching (PPM) compression model to build a character-based model for each author, with works presented in~\cite{neal2017surveying, halvani2018cross};
   \item [BERT] -- combines a BERT pretrained language model with a dense layer for classification, as in~\cite{fabien-etal-2020-bertaa};
   \item [pALM] -- the per-Author Language Model (pALM), also using BERT as described in~\cite{barlas2020cross}.
 \end{description}

 Here, we will focus on the Ngram and BERT classes. 

\subsection{Ngram: AI methods on ROST dataset}
\label{Ngram}
In~\cite{avram2022comparison} we introduced a  dataset, named  \emph{ROmanian Stories and other Texts} (ROST), consisting of Romanian texts that are stories, short stories, fairy tales, novels, articles, and sketches. We collected 400 such texts of different lengths, ranging from 91 to 39195 words. We used multiple AI techniques for classifying the literary texts written by multiple authors by considering a limited number of speech parts (prepositions, adverbs, and conjunctions). The methods we used were Artificial Neural Networks, Support Vector Machines, Multi Expression Programming, Decision Trees with C5.0, and k-Nearest Neighbour.

For the tests performed in~\cite{avram2022comparison} the 400 Romanian texts were divided into training (50\%), validation (25\%), and test (25\%) sets as detailed in Table~\ref{tab:textDiv}. Some of the 5 aforementioned methods required only training and test sets. In such cases, we concatenated the validation set to the training set.

\begin{table}[htbp]
    \small
    \centering
    \caption{List of authors; the number of texts and their distribution on the training, validation, and test sets}
    \label{tab:textDiv}
    \begin{tabular}{c|l|c|c|c|c}
        \hline
         \#&\textbf{Author}&No. of & TrainSet & ValidationSet & TestSet \\ 
         && texts&size&size &size\\ \hline
        0& Ion \textbf{Creangă}& \textbf{28}& 14& 7& 7\\
        1& Barbu Şt. \textbf{Delavrancea}& \textbf{44}& 22& 11& 11\\
        2& Mihai \textbf{Eminescu}& \textbf{27}& 15& 6& 6\\
        3& Nicolae \textbf{Filimon}& \textbf{34}& 18& 8& 8\\
        4& Emil \textbf{Gârleanu}& \textbf{43}& 23& 10& 10\\
        5& Petre \textbf{Ispirescu}& \textbf{40}& 20& 10& 10\\
        6& Mihai \textbf{Oltean}& \textbf{32}& 16& 8& 8\\
        7& Emilia \textbf{Plugaru}& \textbf{40}& 20& 10& 10\\
        8& Liviu \textbf{Rebreanu}& \textbf{60}& 30& 15& 15\\
        9& Ioan \textbf{Slavici}& \textbf{52}& 26& 13& 13\\\hline
        & TOTAL & \textbf{400}	&\textbf{204}&	\textbf{98}&	\textbf{98}\\
    \end{tabular}
    \normalsize
\end{table}

A numerical representation of the dataset was built as vectors of the frequency of occurrence of the considered features. The considered features were inflexible parts of speech (IPoS). Three different sets of IPoS were used. First, only prepositions were considered, then adverbs were added to this list, and finally, conjunctions were added as well. Therefore, three different representations of the dataset (of the 400 texts) were obtained. For each dataset representation (i.e., corresponding to a certain set of IPoS) the numerical vectors 
were shuffled and split into training, validation, and test sets as detailed in Table~\ref{tab:textDiv}. This process (i.e., shuffle and split 50\%–25\%–25\%)  was repeated three times. We, therefore, obtained different dataset representations, which were referred to as described in Table~\ref{tab:datasetsNames}.

\begin{table}[hb]
    \small
    \centering
    \caption{Names used in~\cite{avram2022comparison} to refer to the different dataset representations and their shuffles.  
    \label{tab:datasetsNames}}
    \begin{tabular}{c l l c}
        \hline
        \textbf{\#} & \textbf{Designation} & \textbf{Features to Represent the Dataset} & \textbf{Shuffle}\\\hline
        1 & \textbf{ROST-P-1} & prepositions &  \#1 \\
        2 & \textbf{ROST-P-2} & prepositions& \#2 \\
        3 & \textbf{ROST-P-3} & prepositions& \#3 \\\hline
        4 & \textbf{ROST-PA-1} & prepositions and adverbs & \#1 \\
        5 & \textbf{ROST-PA-2} &  prepositions and adverbs & \#2 \\
        6 & \textbf{ROST-PA-3} &  prepositions and adverbs & \#3 \\\hline
        7 & \textbf{ROST-PAC-1} &  prepositions, adverbs and conjunctions & \#1 \\
        8 & \textbf{ROST-PAC-2} &  prepositions, adverbs and conjunctions & \#2 \\
        9 & \textbf{ROST-PAC-3} &  prepositions, adverbs and conjunctions & \#3 \\\hline
        \hline
    \end{tabular}
    \normalsize
\end{table}

 All these representations of the dataset as vectors of the frequency of occurrence of the considered feature lists can be found as text files at reference~\cite{sanda2022kaggle}. These files contain feature-based numerical value representations for a different text on each line. On the last column of 
these files are numbers from 0 to 9 corresponding to the author, as specified in the first columns of Table~\ref{tab:textDiv}.

\begin{figure}[htbp]
    \centering
    \includegraphics[width=.62\textwidth]{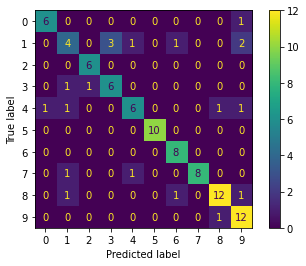}
    \caption{Confusion matrix on MEP's best results. The numbers from 0 to 9 are the codes given to our authors, as specified in the first column of Table~\ref{tab:textDiv}.}
    \label{fig:confMatMEPX}
\end{figure}

The best value obtained amongst all aforementioned AI methods was on ROST-PA-2 by using MEP, with a test error rate of 20.40\%. 
This means that 20 out of 98 tests were misclassified. The obtained the \emph{Confusion Matrix} depicted in Figure~\ref{fig:confMatMEPX}. The \emph{macro-accuracy} value obtained was 80.94\%. The Python code for building the \emph{Confusion Matrix} and calculating the  \emph{macro-accuracy} value is provided at~\cite{avram2023macroAccuracy}.

\subsection{BERT}
Bidirectional Encoder Representations from Transformers (BERT) is a language representation model designed to pretrain deep bidirectional representations from the unlabeled text by jointly conditioning on both left and right context in all layers~\cite{devlin2018bert}. It was first introduced in October 2018 by a team from Google, being now used as a machine learning framework for natural language processing (NLP) tasks  (i.e., classification, entity recognition, question answering, etc.).

BERT is based on multiple techniques that preceded it, starting from \emph{RNN} and \emph{LSTMs} to \emph{Attention} and \emph{Transformers}. Starting from 2015, \emph{Recurrent Neural Networks} (RNNs) were used for author identification~\cite{bagnall2015author}. However, RNNs have a problem with remembering long-term
dependencies (over ten sequences)~\cite{zhang2018developing}. \emph{Long Short Term Memory} (LSTM) architecture~\cite{hochreiter1997long} is a special kind of RNN, introduced to overcome this weakness of the traditional RNN. 

\emph{RNN encoder-decoder}~\cite{sutskever2014sequence} architectures were used mostly for translations. In this context, the encoder translates the input into a ``codified'' vector, which is then passed to the decoder, which translates it again into the output language. The input and output vectors do not need to have the same length. The challenge with the encoder-decoder approach is that it has to transform the input sentence into a fixed-length vector that becomes a bottleneck. Therefore, such a model has difficulty translating long sentences. The \emph{attention} technique was introduced in~\cite{bahdanau2014neural,luong2015effective}, through which instead of passing the last hidden state of the encoding stage, the encoder passes all the hidden states to the decoder. This allows access to parts of the source sentence that are relevant to predicting a target word. In 2017, the concept of \emph{Multi-Head Self Attention} was introduced in the paper ``Attention Is All You Need''~\cite{vaswani2017attention}. This technique is called \emph{transformers} and uses self-attention for language understanding,  parallelization for better translation quality, and attention to integrate the relevance of a set of values (information) based on some keys and queries. Google proposed BERT~\cite{devlin2018bert} in 2018, which uses a bi-directional transformer. Due to its complexity, more data is required for training such models. Therefore, there are currently many pretrained models in different languages. 

BERT was used for AA in~\cite{manolache2021transferring}, while the performance of BERT was compared with other techniques~\cite{bagnall2015author, barlas2020cross, tyo2022state} for solving the AA problem. Work in~\cite{fabien-etal-2020-bertaa} concludes that BERT is the highest-performing AA method.

\subsection{Comparing methods}
The multitude of AA approaches makes this problem difficult to have a unified view of the state-of-the-art results. In~\cite{tyo2022state}, authors highlight this challenge by noting significant differences in:
\begin{description}
  \item \textbf{Datasets}
  \begin{itemize}
    \item  size: small (CCAT50, CMCC, Guardian10), medium (IMDb62, Blogs50), and large (PAN20, Gutenberg);
    \item  content: cross-topic ($\times_t$), cross-genre ($\times_g$);
    \item  imbalance (imb): i.e., the standard deviation of the
number of documents per author; 
    \item  topic confusion (as detailed in~\cite{altakrori-etal-2021-topic-confusion}).
  \end{itemize}
  \item \textbf{Performance metrics}
  \begin{itemize}
    \item  type: accuracy, F1, c@1, recall, precision, macro-accuracy, AUC, R@8, and others;
    \item computation: even for the same performance metrics there were different ways of computing them.
  \end{itemize}
  \item \textbf{Methods}
  \begin{itemize}
    \item  the feature extraction method:
    \begin{itemize}
      \item Feature-based: n-gram, summary statistics, co-occurrence graphs;
      \item Embedding-based: char or word embedding, transformers;
      \item Feature and embedding-based: BERT.
    \end{itemize}
  \end{itemize}
\end{description}

The work presented in~\cite{tyo2022state} tries to address and ``resolve'' these differences, bringing everything to a common denominator by using \emph{macro-accuracy} as metric.

The overall accuracy, also known as \emph{micro-accuracy} or \emph{micro-averaged accuracy} weights the accuracy for each sample (text) equally. The \emph{macro-accuracy} metric, also known as \emph{macro-averaged accuracy} is regarded to be more accurate in the case of imbalanced datasets, as it is computed by weighting equally the accuracy for each class (author).

A formal description of computing micro- and macro-accuracy was provided by Jacob Tyo in~\cite{2023Tyo_pers} and is described next:

\begin{align*}
N &=\text{total number of texts}\\
M &=\text{number of authors}\\
N_i &=\text{number of texts for author }i\\
C_i &=\text{number of texts correctly predicted for author }i\\
\mathcal{A}_{micro} &= \text{micro-averaged accuracy} \\
\mathcal{A}_{macro} &= \text{macro-averaged accuracy} \\
&\\ 
\mathcal{A}_{micro} &= {1 \over N} \sum_{i=1}^{N} C_{i} \\
\mathcal{A}_{macro} &= {1 \over M} \sum_{i=1}^{M} {C_i  \over N_i} 
\end{align*}



The results of the state of the art as presented in~\cite{avram2022comparison} are shown in Table~\ref{tab:StateOfArt}.

\begin{landscape}
\begin{table}[p]
    \caption{State of the art \emph{macro-accuracy} of authorship attribution models. Information collected from~\cite{tyo2022state} (Tables 1 and 3). 
    \emph{Name} is the name of the dataset; \emph{No. docs} represents the number of documents in that dataset; \emph{No. auth} represents the number of authors; \emph{Content} indicates whether the documents are cross-topic ($\times_t$) or cross-genre ($\times_g$); \emph{W/D} stands for \emph{words per document}, representing the average length of documents; \emph{imb} represents the \emph{imbalance} of the dataset 
    measured by the standard deviation of the number of documents per author.\label{tab:StateOfArt}}
    \begin{tabular}{c|c|c|c|c|c|c|c|c|c}
        \hline
        \multicolumn{6}{c}{\textbf{Dataset }} & \multicolumn{4}{c}{\textbf{Investigation Type} }\\\hline
        \textbf{Name} & \textbf{No. Docs} & \textbf{No. Auth} & \textbf{Content}& \textbf{W/D}& \textbf{Imb}& \textbf{Ngram} & \textbf{PPM} & \textbf{BERT} & \textbf{pALM}\\\hline
        CCAT50 & 5000&50& - & 506& 0& 76.68& 69.36 & 65.72 & 63.36\\
        CMCC & 756&21& $\times_t$ $\times_g$&601& 0& 86.51 & 62.30 & 60.32 & 54.76\\
        Guardian10& 444&13 & $\times_t$ $\times_g$& 1052& 6.7& 100 & 86.28 & 84.23 & 66.67 \\
            ROST & 400&10& $\times_t$ $\times_g$&3355& 10.45 &80.94& $-$ & $-$ & $-$ \\
        IMDb62 & 62,000&62 & -&349& 2.6& 98.81 & 95.90 & 98.80 & -\\
        Blogs50 & 66,000&50&- &122& 553& 72.28& 72.16 & 74.95 & - \\
        PAN20 & 443,000&278,000 &$\times_t$ &3922&2.3& 43.52 & - & 23.83 & -\\
        Gutenburg & 29,000&4500& -&66,350 &10.5& 57.69& - & 59.11 & - \\
    \hline
    \end{tabular}
\end{table}
\end{landscape}

For the work presented in~\cite{avram2022comparison} we did not investigate how BERT would perform on ROST. Therefore we will conduct this investigation here.

\section{Prerequisite}
\label{prerequisite}
NLP models such as those presented in Section~\ref{Ngram}, and LSTMs or CNNs require inputs in the form of numerical vectors, and this typically means translating features like the vocabulary and parts of speech into numerical representations. In contrast with other solutions that produce such numerical representations by word frequencies or uniquely indexed values, BERT produces word representations that are dynamically informed by the words around them~\cite{mccormick2019BERTtutorial}, meaning that the same word would have a different numerical representation based on the context and meaning in that context.

BERT is a pretrained model that expects input data in a specific format~\cite{mccormick2019BERTtutorial}. 
The pretrained model creates a fixed-size vocabulary of individual characters, subwords, and words that best fits the data. Then, the tokenizer first checks if the whole word is in the vocabulary. If not, it tries to break the word into the largest possible subwords contained in the vocabulary, and as a last resort will decompose the word into individual characters~\cite{wu2016google}. This process increases the length of the initial input vector.  

BERT is able to process vectors of maximum of 512 tokens, out of which the start token, i.e., [CLS], and the end token, i.e., [SEP], are predefined. Both these tokes are always required. 

Considering all these requirements, we needed to adjust the size of our texts to be proper inputs for the BERT model. By running different preliminary tests, where we tried lengths of 91, 200, and 400 words, we obtained better results for texts of 200 words. The 91 value is the number of words of the shortest text among the 400 texts in our dataset. The length of these texts ranges from 91 to 39195 words. 

We have set the maximum length of the input vector to 256 to accommodate the situations where some words are not in the pretrained model vocabulary and therefore, may be divided into subwords or even individual characters. Considering that some of the texts are written a couple of centuries ago, that might happen especially for words that are no longer used. 

We kept the initial shuffles as described in Table~\ref{tab:textDiv} with the following changes:
\begin{itemize}
    \item for each text that was longer than 200 words we divided it into multiple texts of 200 words maximum;
    \item we have only training and test sets, the validation sets being added to the training sets.
\end{itemize}

Thus, we obtained 3 shuffles of the dataset with the number of texts as detailed in Table~\ref{tab:textDiv4BERT}.

\begin{table}[htbp]
    \small
    \centering
    \caption{List of authors; the number of texts and their distribution on the training and test sets prepared for BERT}
    \label{tab:textDiv4BERT}
    \begin{tabular}{c|l|c|c|c}
        \hline
         \#&\textbf{Author}&No. of & TrainSet size & TestSet size\\ 
         && texts&shuffle 1, 2, 3 &shuffle 1, 2, 3\\ \hline
        0& Ion \textbf{Creangă}& \textbf{520}& 446,399,393& 74,121,127\\
        1& Barbu Şt. \textbf{Delavrancea}& \textbf{922}&706,595,690& 216,327,232\\
        2& Mihai \textbf{Eminescu}& \textbf{792}& 550,467,447& 242,325,345\\
        3& Nicolae \textbf{Filimon}& \textbf{475}& 389,326,278& 86,149,197\\
        4& Emil \textbf{Gârleanu}& \textbf{203}& 169,146,160& 34,57,43\\
        5& Petre \textbf{Ispirescu}& \textbf{674}& 489,469,482& 185,205,192\\
        6& Mihai \textbf{Oltean}& \textbf{106}& 81,74,86& 25,32,20\\
        7& Emilia \textbf{Plugaru}& \textbf{463}& 386,329,312& 77,134,151\\
        8& Liviu \textbf{Rebreanu}& \textbf{711}& 535,523,546& 176,188,165\\
        9& Ioan \textbf{Slavici}& \textbf{1966}& 1660,1467,1445& 306,499,521\\\hline
        & TOTAL & \textbf{6832}	&\textbf{5411,4795,4839}&	\textbf{1421,2037,1993}\\
    \end{tabular}
    \normalsize
\end{table}

\section{Tests}
\label{tests}

For our tests, we used Google Colab Pro\cite{bisong2019google},  an online framework that offers different runtime environments to run code written in Python.

\subsection{Fine-tuning a pretrained Romanian BERT model}
\label{finetune}
We used the Romanian pretrained BERT model called ``dumitrescustefan/bert-base-romanian-cased-v1'', which is described in~\cite{dumitrescu-etal-2020-birth} and available at~\cite{bert-base-romanian}. 

For training and testing, we used the following classes from HuggingFace's Transformers~\cite{wolf-etal-2020-transformers} library: 
\begin{description}
    \item [ AutoTokenizer] - to download the tokenizer associated to the pretrained Romanian model we chose;
    \item [ AutoModelForSequenceClassification] - to download the model itself;
    \item [ Trainer] and \textbf{TrainingArguments} - to fine-tune the chosen model for our dataset, namely ROST.
\end{description}

 The preprocessing of the texts done by the tokenizer consists in splitting the input text into words (or parts of words, punctuation symbols, etc.) usually called tokens, which then are converted to ids/numbers. In order to process all the input texts as a batch, one needs to pad them all to the same length or truncate them to the maximum length by specifying that to the tokenizer. 

 The configuration for our tests is detailed in Table~\ref{tab:config}.

 \begin{table}[htbp]
    \centering
    \small
    \caption{Configuration parameters}
    \label{tab:config}   
    \begin{tabular}{ l l }
        \hline 
        \textbf{Parameter} & \textbf{Value}\\
        \hline
          architectures & BertForSequenceClassification \\
          hidden\_act & gelu \\
          hidden\_dropout\_prob & 0.1\\
          hidden\_size & 768\\
          initializer\_range & 0.02 \\
          intermediate\_size & 3072\\
          layer\_norm\_eps & 1e-12 \\
          num\_attention\_heads & 12 \\ 
          num\_hidden\_layers & 12\\
          torch\_dtype & float32\\
          transformers\_version & 4.26.0\\
          vocab\_size & 50000 \\
          \hline
    \end{tabular}
 \normalsize
\end{table}

For our training, the parameters used are presented in Table~\ref{tab:training_settings}.

\begin{table}[htbp]
    \centering
    \small
    \caption{BERT on ROST parameters}
    \label{tab:training_settings}   
    \begin{tabular}{ l l }
        \hline 
        \textbf{Parameter} & \textbf{Value}\\
        \hline
          Evaluation strategy & epoch \\
          Num Epochs & 3 \\
          Instantaneous batch size per device & 8\\
          Total train batch size (w. parallel, distributed, and accumulation) & 8\\
          Test Batch size & 8 \\
          Gradient Accumulation steps & 1\\
          Learning rate & 5e-05 \\
          Seed & 42 \\ 
          Total optimization steps & 2031, 1800, 1815\\
          Number of trainable parameters & 124449034\\
          Tokenizer maximum length & 256\\
          \hline
    \end{tabular}
    \normalsize
\end{table}

There are three different values for \emph{Total optimization steps} corresponding to the three shuffles.

Some training performance metrics we obtained are presented in Table~\ref{tab:training_perf}.

\begin{table}[htbp]
    \centering
    \small
    \caption{BERT on ROST training performance metrics}
    \label{tab:training_perf}   
    \begin{tabular}{ l l l l }
        \hline 
        \textbf{Parameter} & \textbf{for shuffle 1} & \textbf{for shuffle 2} & \textbf{for shuffle 3}\\
        \hline
        train\_runtime (in seconds) & 261.1692 & 239.6519 & 241.3783\\
        train\_samples\_per\_second& 62.155 & 60.025 & 60.142\\
        train\_steps\_per\_second& 7.777 &  7.511 &7.519\\
        train\_loss& 0.255753& 0.297084 & 0.381490\\
     \hline
    \end{tabular}
    \normalsize
\end{table}

\subsection{Evaluation}
\label{evaluation}

For evaluating the results we used HuggingFace's Evaluate~\cite{wolf-etal-2020-transformers}. With this library, we computed the \emph{micro-accuracy} metric (aka. \emph{accuracy} or \emph{micro-averaged accuracy}). The corresponding values for this metric, obtained for the three shuffles are shown in Table~\ref{tab:eval_perf} alongside other evaluation performance metrics.

 \begin{table}[htbp]
    \centering
    \small
    \caption{BERT on ROST evaluation performance metrics}
    \label{tab:eval_perf}   
    \begin{tabular}{ l l l l }
        \hline 
        \textbf{Parameter} & \textbf{for shuffle 1} & \textbf{for shuffle 2} & \textbf{for shuffle 3}\\
        \hline 
        eval\_accuracy&0.864883 &0.864997&0.849974\\
        eval\_runtime&6.0036 &8.474&8.299\\
        eval\_samples\_per\_second&236.69 &240.382&240.149\\
        eval\_steps\_per\_second & 29.649 &30.092&30.124\\
          \hline
    \end{tabular}
    \normalsize
\end{table}

\begin{figure}[htbp]
    \centering
    \includegraphics[width=.61\textwidth]{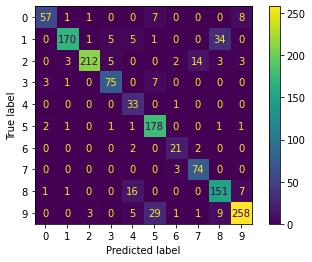}
    \includegraphics[width=.61\textwidth]{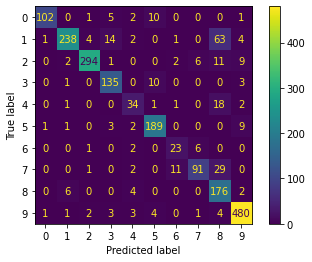}
    \includegraphics[width=.61\textwidth]{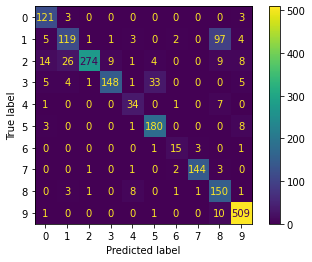}

    \caption{Confusion matrices for (a) shuffle 1 (b) shuffle 2 (c) shuffle 3} 
        \label{fig:CMs}
\end{figure}

We also used the sklearn.metrics module from the Scikit-learn~\cite{scikit-learn} library. The classes used, pertaining to this module, were: 
\begin{description}
    \item [ confusion\_matrix] - to compute the confusion matrix;
    \item [ ConfusionMatrixDisplay] - to generate a Confusion Matrix visualization as the ones displayed in Figures~\ref{fig:CMs};
    \item [ classification\_report] - to obtain a report with main classification metrics;
    \item [ accuracy\_score] - to compute the accuracy classification score, also known as micro-accuracy or overall accuracy;
    \item [ balanced\_accuracy\_score] - to compute the balanced accuracy, also known as macro-accuracy.
\end{description}

To compute the confusion matrix, we provided the true and predicted classes (i.e., codes corresponding to the authors). The graphical visualizations of the confusion matrices corresponding to the results obtained for the 3 shuffles are depicted in Figure~\ref{fig:CMs}. The numbers from 0 to 9 are the codes given to our authors, as specified in the first column of Table~\ref{tab:textDiv4BERT}.

Figure~\ref{fig:CMs} shows a significant confusion reoccurring between Barbu Șt. 
Delavrancea (1) and Liviu Rebreanu (8). The texts written by these authors occurred in almost the same time period (i.e., 1884–1909 and 1908–1935 respectively). 

For each class/author, the classification report provided various metrics:
\begin{description}
  \item \emph{Precision}---the number of correctly attributed authors divided by the number of instances when the algorithm identified the attribution as correct;
  \item \emph{Recall (Sensitivity)}---the number of correctly attributed authors divided by the number of test texts belonging to that author;
  \item \emph{F1-score}---a weighted harmonic mean of the \emph{Precision} and \emph{Recall}.
\end{description}

The classification results for the 3 shuffles are shown in Tables~\ref{tab:ClassificatioReport-shuffle1}, ~\ref{tab:ClassificatioReport-shuffle2} and~\ref{tab:ClassificatioReport-shuffle3}.

\begin{table}[htbp] 
    \small
    \caption{Shuffle 1 results generated by sklearn.metrics's classification\_report
    \label{tab:ClassificatioReport-shuffle1}}
    \begin{tabular}{r r c c c l}
         \hline
         \textbf{Name}  & class&\textbf{precision}  &  \textbf{recall} & \textbf{f1-score} &  support \\\hline 
        Ion Creangă&  0 & 0.905 & 0.770 & 0.832 & 74 \\
        Barbu Șt. Delavrancea  & 1 & 0.960 & 0.787 & 0.865 & 216 \\
        Mihai Eminescu  & 2 & 0.977 & 0.876 & 0.924 & 242 \\
        Nicolae Filimon  & 3 & 0.872 & 0.872 & 0.872 & 86 \\
        Emil Gârleanu  & 4 & 0.532 & 0.971 & 0.688 & 34 \\
        Petre Ispirescu  & 5 & 0.802 & 0.962 & 0.875 & 185 \\
        Mihai Oltean  & 6 & 0.750 & 0.840 & 0.792 & 25 \\
        Emilia Plugaru  & 7 & 0.813 & 0.961 & 0.881 & 77 \\
        Liviu Rebreanu  & 8 & 0.763 & 0.858 & 0.807 & 176 \\
        Ioan Slavici  & 9 & 0.931 & 0.843 & 0.885 & 306 \\
        \hline 
        &\textbf{accuracy} &  &   &     0.865 & 1421 \\
        &\textbf{macro avg}  & 0.831 & \textbf{0.874} & 0.842 & 1421 \\
        &\textbf{weighted avg} & 0.882 & 0.865 & 0.868 & 1421 \\
        \hline
    \end{tabular}
    \normalsize
 \end{table}

\begin{table}[htbp] 
    \small
    \caption{Shuffle 2 results generated by sklearn.metrics's classification\_report
    \label{tab:ClassificatioReport-shuffle2}}
    \begin{tabular}{r r c c c l}
         \hline
         \textbf{Name}  & class&\textbf{precision}  &  \textbf{recall} & \textbf{f1-score} &  support \\\hline 
        Ion Creangă & 0 & 0.971 & 0.843 & 0.903 &  121\\
        Barbu Șt. Delavrancea &  1 & 0.952 & 0.728 & 0.825 &  327\\
        Mihai Eminescu  & 2 & 0.970 & 0.905 & 0.936 &  325\\
        Nicolae Filimon &  3 & 0.839 & 0.906 & 0.871 &  149\\
        Emil Gârleanu  & 4 & 0.667 & 0.596 & 0.630 &   57\\
        Petre Ispirescu &  5 & 0.883 & 0.922 & 0.902 &  205\\
        Mihai Oltean &  6 & 0.605 & 0.719 & 0.657 &   32\\
        Emilia Plugaru &  7 & 0.875 & 0.679 & 0.765 &  134\\
        Liviu Rebreanu &  8 & 0.585 & 0.936 & 0.720 &  188\\
        Ioan Slavici &  9 & 0.941 & 0.962 & 0.951 &  499\\
        \hline 
        &\textbf{accuracy}   &  &  &   0.865 & 2037\\
        &\textbf{macro avg}  & 0.829 & \textbf{0.820} & 0.816 & 2037\\
        &\textbf{weighted avg}  & 0.886 & 0.865 & 0.868 & 2037\\
    \hline
    \end{tabular}
    \normalsize
 \end{table}
 
\begin{table}[htbp] 
     \small
     \caption{Shuffle 3 results generated by sklearn.metrics's classification\_report
     \label{tab:ClassificatioReport-shuffle3}}
     \begin{tabular}{r r c c c l}
         \hline
         \textbf{Name}  & class&\textbf{precision}  &  \textbf{recall} & \textbf{f1-score} &  support \\\hline 
         Ion Creangă& 0 & 0.807 & 0.953 & 0.874 &  127 \\
        Barbu Șt. Delavrancea &  1 & 0.768 & 0.513 & 0.615 &  232 \\
        Mihai Eminescu  & 2 & 0.986 & 0.794 & 0.880 &  345 \\
        Nicolae Filimon &  3 & 0.937 & 0.751 & 0.834 &  197 \\
        Emil Gârleanu &  4 & 0.694 & 0.791 & 0.739 &   43 \\
        Petre Ispirescu &  5 & 0.822 & 0.938 & 0.876 &  192 \\
        Mihai Oltean &  6 & 0.714 & 0.750 & 0.732 &   20 \\
        Emilia Plugaru &  7 & 0.973 & 0.954 & 0.963 &  151 \\
        Liviu Rebreanu  & 8 & 0.543 & 0.909 & 0.680 &  165 \\
        Ioan Slavici  & 9 & 0.944 & 0.977 & 0.960 &  521 \\
        \hline 
        &\textbf{accuracy}   &  &  &   0.850 & 1993\\
        &\textbf{macro avg} & 0.819 & \textbf{0.833} & 0.815 & 1993 \\
        &\textbf{weighted avg} & 0.871 & 0.850 & 0.850 & 1993 \\
        \hline
    \end{tabular}
    \normalsize
 \end{table}

\begin{figure}[h]
    \centering
    \includegraphics[width=.47\textwidth]{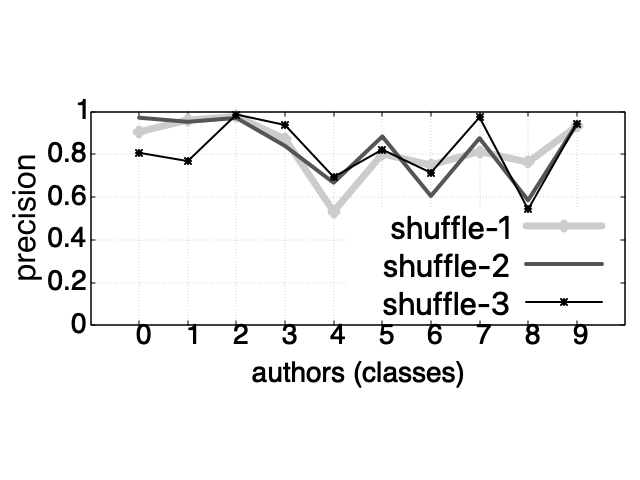}
    \includegraphics[width=.47\textwidth]{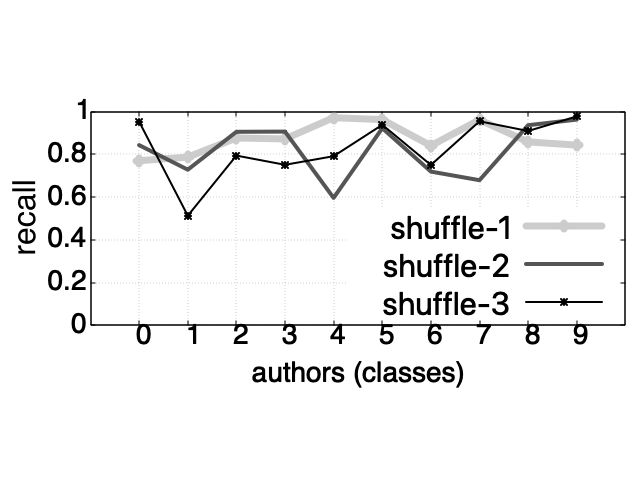}
    \includegraphics[width=.47\textwidth]{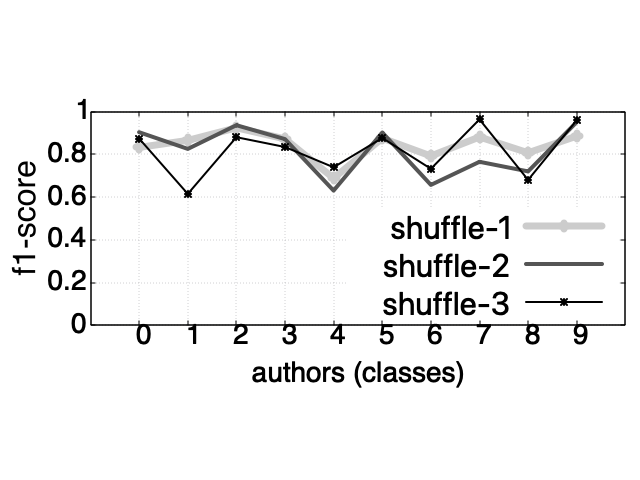}

    \caption{Graphical representations of results from Tables~\ref{tab:ClassificatioReport-shuffle1}-\ref{tab:ClassificatioReport-shuffle3} for (a) precision (b) recall (c) f1-score} 
        \label{fig:grRez}
\end{figure}

 A graphical representation of the results obtained per classes and detailed in Tables~\ref{tab:ClassificatioReport-shuffle1}, ~\ref{tab:ClassificatioReport-shuffle2} and~\ref{tab:ClassificatioReport-shuffle3} are presented in Figure~\ref{fig:grRez}.

As can be seen, the worst results over the three shuffles are obtained consistently by Emil Gârleanu (4), Mihai Oltean (6), and Liviu Rebreanu (8), while the best results are obtained by Mihai Eminescu (2) and Emilia Plugaru (7) and Petre Ispirescu (5). 

Table~\ref{tab:accu} presents the best micro- and macro-accuracies obtained on the three shuffles.
\begin{table}[htbp] 
    \centering
     \small
     \caption{Micro- and macro-accuracies 
     \label{tab:accu}}
     \begin{tabular}{ l l l l }
        \hline 
        \textbf{accuracy} & \textbf{for shuffle 1} & \textbf{for shuffle 2} & \textbf{for shuffle 3}\\
        \hline 
        micro-accuracy&0.864883 &0.864997&0.849974\\
        macro-accuracy&0.874031 &0.819585&0.832905\\
        \hline
    \end{tabular}
    \normalsize
 \end{table}

The best overall accuracy is 0.864997, and it is obtained on the second shuffle. That is interesting as in the investigations performed in~\cite{avram2022comparison} by using other AI techniques, also the second shuffle obtained the best results. This metric is also referred to as \emph{micro-accuracy}, treating each sample (or, in this case, each text fragment) as equally important. 

A more representative metric is the \emph{macro-accuracy} as it computes the accuracy by treating each author equally. The best macro-accuracy obtained is 0.874031, and it is obtained on the first shuffle.

\section{Conclusion and further work}
\label{conclusion}
In this paper, we continued our investigation started in~\cite{avram2022comparison}, by addressing the authorship attribution problem on a Romanian dataset by using BERT. For that, we needed to use shorter texts (200 tokens/words). This lead to a more balanced dataset than the one used in~\cite{avram2022comparison} as the texts have approximately the same length. However, the other perspectives, such as the  number of texts per author, the sources from which the texts were collected, the time period in which the authors lived and wrote these texts, the intended reading medium (i.e., paper or online), and the type of writing (i.e., stories, short stories, fairy tales, novels, literary articles, and sketches), remained. 

The best result obtained in~\cite{avram2022comparison} was 80.94\%. The best results obtained here with BERT are 85.90\%. However, we cannot compare the two as the dataset was modified to accommodate BERT's processing capabilities. To be able to compare all these methods, one needs to recreate the experiments made with  Artificial Neural Networks, Support Vector Machines, Multi Expression Programming, Decision Trees with C5.0, and k-Nearest Neighbour by using the 200 tokens/words texts.  

As we conducted these first BERT tests, we divided our texts into parts of 200 words disregarding the sentence integrity. We would like to conduct some tests to see if keeping the integrity of sentences would have an impact on our results. 

As a future direction, we would also like to explore more the Prediction by Partial Matching (PPM) mentioned in Section~\ref{related_work}.

\label{future_work}

\bibliographystyle{unsrt}
\bibliography{main}
\end{document}